\documentclass{article}
\usepackage{spconf,amsmath,graphicx}
\usepackage{url}
\usepackage{enumitem}

\title{Word Embeddings from Large-Scale Greek Web Content}
\name{\parbox{\linewidth}{\centering Stamatis Outsios$^1$, Konstantinos Skianis$^2$, Polykarpos Meladianos$^1$, \\ \it Christos Xypolopoulos$^2$, Michalis Vazirgiannis$^{1,2}$}}
\address{
\centering
\parbox{\columnwidth}{\centering
$^1$ Athens University of Economics and Business \\
	Department of Informatics\\
	Greece}
\parbox{\columnwidth}{\centering
	$^2$ \'Ecole Polytechnique\\
	Laboratoire d'informatique (LIX)\\
	France}
}

\begin{document}
%
\maketitle
\begin{abstract}
Word embeddings are undoubtedly very useful components in many NLP tasks.
In this paper, we present word embeddings and other linguistic resources trained on the largest to date digital Greek language corpus.
We also present a live web tool for testing the Greek word embeddings\footnote{\url{http://archive.aueb.gr:7000}\label{demo-url}}, by offering ``analogy", ``similarity score" and ``most similar words" functions.
Through our explorer, one could interact with the Greek word vectors.
\end{abstract}
\section{Introduction \& Related work}
\label{sec:intro}

With the rise of neural networks and deep learning in the NLP community \cite{bengio2003neural,collobert2008unified}, word embeddings were introduced \cite{mikolov2013distributed}, having a huge impact on numerous tasks.
Their ability to represent rich relationships between words, led to state-of-the-art results, combined with CNNs \cite{kim2014convolutional} and LSTMs \cite{Johnson:2016} in text classification, question answering and machine translation tasks.
Although word vector transformation is the most common way to harvest text, they require a large amount of data in order to be trained.
Moreover, resources for specific languages may be scarce or hard to extract in an efficient way.

In this work, we present Greek word embeddings, trained on, to the best of our knowledge, the largest so far corpus available,  collected/crawled from about 20M URLs with Greek language content. 
The vocabulary and word vectors are available on request.
We developed a live web tool to enable users to interact with the Greek word embeddings.
Some of the functions we provide is similarity score, most similar words as well as analogy operations.
We also present a vector explorer, where we can project a sample of the word vectors.


Lately, pre-trained word vectors for 294 languages were introduced, trained on Wikipedia using FastText\cite{bojanowski2017enriching}.
These 300-dimensional vectors were obtained using the skip-gram model.
Their Greek language variant, was trained on the Wikipedia corpus concerning only Greek documents.

Visualization tools for word embeddings are of great importance, since they contribute to the interpretation of their nature.
Similarly to our tool, Tensorflow\footnote{\url{https://projector.tensorflow.org/}} offers an illustration of a sample of word embeddings after applying dimensionality reduction techniques.
Last, the training process can be observed with \texttt{WEVI}, a word embedding visual inspector\footnote{\url{https://ronxin.github.io/wevi/}}.

\section{Crawling the Greek Web}
\label{sec:pagestyle}

For the process of crawling the Greek Web (which was funded by the Stavros Niarhhos foundation, see: https://www.snf.org, for the Greek National Library) we used the \texttt{Heritrix}\footnote{\url{http://crawler.archive.org/}} tool.
Collecting the websites adheres to the international Web Archive (WARC) template. The WARC file form defines a method combining multiple media resources into one archive. 
Next, we present some statistics about the data we crawled:
\begin{itemize}[noitemsep]
\item Number of WARCs: 112K
\item Size of HTML (stored in WARC format): 10TB
\item Number of Greek domains: 350K
\item Number of URLs: 20M
\item Duration of crawling: 45 days
\end{itemize}

\section{PRE-PROCESSING \& TEXT EXTRACTION}
Before training, we applied several pre-processing and extraction steps on the raw crawled text:
\begin{enumerate}[noitemsep]
\item detect the encoding of each webpage, so that we are able to read it properly,
\item remove HTML code and tags, as well as Javascript,
\item remove boilerplate code\footnote{\url{https://en.wikipedia.org/wiki/Boilerplate_code}},
\item remove all non-Greek characters,
\item track the line change character,
\item produce compressed text files per domain.
\end{enumerate}
The third step is very important for the text quality, since we request a corpus that can be used later for developing linguistic resources (language model, embeddings etc.).
Except their content, webpages consist of navigation elements, headers, footers, as well as commercial banners.
This text is usually not associated with the webpage's main content, and can lead in decreasing the integrity of the collection.
In order to do that, we used libraries like \texttt{BeautifulSoup}, \texttt{Justext}, \texttt{NTLK}’s \texttt{clean\_html} and \texttt{Boilerpipe}\footnote{\url{https://boilerpipe-web.appspot.com/}}.
The best results were obtained by \texttt{Boilerpipe}, which was the one we used in the end to remove useless text (boilerplate).
We removed identical sentences (de-duplication) and produced the final corpus in text form, sized around 50GB.
We obtained thus   $\sim$3B tokens and a total number of 498M sentences, with 118M of them being unique.

De-duplication per domain resulted in reducing the size of raw corpus by 75\%.
With an additional processing of the final corpus, we create the Greek language n-grams: Unigrams: $\sim$7M, Bigrams: $\sim$90M, Trigrams: $\sim$300M.

\section{Training Greek Word Embeddings}
\label{sec:training}

For the process of learning the Greek word embeddings we utilized the FastText\cite{bojanowski2017enriching} library, which takes under consideration the morphology of a word.
Training on the raw uncompressed text of the Greek internet web, with size of 50GB, required 2 days in a 8-core Ubuntu system with 32GB of RAM.

Different Greek vector models were produced like:
1. native fasttext skipgram with the following parameters:
-minCount 11 -loss ns -thread 8 -dim 300,
2. native fasttext cbow,
3. gensim\footnote{\url{https://radimrehurek.com/gensim/}} word2vec skipgram,
4. gensim fasttext skipgram,
5. gensim fasttext skipgram, no subword information.
Methods 3 and 5 lead to the same result as they use the same technique.
By evaluating their effectiveness in automatic spell correction along with similarity queries, method 1 yields the most reliable results.
In the future, we plan to offer as well a set of handcrafted questions for evaluation purposes. 

\section{Visualization}
\label{sec:demo}
Next, we designed tools in order to visualize examples of Greek word vector relationships.
The first demo offers linguistic functions which are enabled by the existence of word embeddings, like analogy, similarity score or most similar words.
The second demo tool for exploring and querying the word vectors was based on the \texttt{word2vec-explorer}\footnote{\url{https://github.com/dominiek/word2vec-explorer}}.
In this tool, a user can navigate through a sample of the Greek word embeddings, visualize it via t-SNE\cite{maaten2008visualizing} and apply k-means clustering.
Furthermore, we offer comparing functions for combinations of words.
For the frontend, we used libraries like \texttt{Flask}, \texttt{Jinja} and \texttt{Bootstrap}.

\section{Conclusion \& Future Work}
\label{sec:conclusion}
In this work, we present the efforts that resulted in Greek word embeddings and other Greek language resources trained on the largest corpus available for the Greek language.
The resources (corpus, trained vectors, stopwords, vocabulary as well as unigrams, bigrams and trigrams) are available on request.
We have also implemented a live web tool, where a user can explore word relationships in the Greek language.
In addition, we provide a word embedding explorer, where one could visualize a sample of the Greek vectors with \texttt{t-SNE}\cite{maaten2008visualizing}.

Recently, embeddings evolved towards new approaches like Hierarchical Representations \cite{NIPS2017_7213} or ELMo \cite{Peters:2018}.
Finally, we plan to extend our work by adding a visualization of the Word Mover's Distance\cite{kusner2015doc}.


\bibliographystyle{IEEEbib}
\bibliography{refs}

\begin{thebibliography}{10}

\bibitem{bengio2003neural}
Yoshua Bengio, R{\'e}jean Ducharme, Pascal Vincent, and Christian Jauvin,
\newblock ``A neural probabilistic language model,''
\newblock {\em JMLR}, 2003.

\bibitem{collobert2008unified}
Ronan Collobert and Jason Weston,
\newblock ``A unified architecture for natural language processing: Deep neural
  networks with multitask learning,''
\newblock in {\em ICML}, 2008.

\bibitem{mikolov2013distributed}
Tomas Mikolov, Ilya Sutskever, Kai Chen, Greg~S Corrado, and Jeff Dean,
\newblock ``Distributed representations of words and phrases and their
  compositionality,''
\newblock in {\em NIPS}, 2013, pp. 3111--3119.

\bibitem{kim2014convolutional}
Yoon Kim,
\newblock ``Convolutional neural networks for sentence classification,''
\newblock in {\em EMNLP}, 2014.

\bibitem{Johnson:2016}
Rie Johnson and Tong Zhang,
\newblock ``Supervised and semi-supervised text categorization using lstm for
  region embeddings,''
\newblock in {\em ICML}, 2016.

\bibitem{bojanowski2017enriching}
Piotr Bojanowski, Edouard Grave, Armand Joulin, and Tomas Mikolov,
\newblock ``Enriching word vectors with subword information,''
\newblock {\em TACL}, 2017.

\bibitem{maaten2008visualizing}
Laurens van~der Maaten and Geoffrey Hinton,
\newblock ``Visualizing data using t-sne,''
\newblock {\em JMLR}, 2008.

\bibitem{NIPS2017_7213}
Maximillian Nickel and Douwe Kiela,
\newblock ``Poincar\'{e} embeddings for learning hierarchical
  representations,''
\newblock in {\em NIPS}. 2017.

\bibitem{Peters:2018}
Matthew~E. Peters, Mark Neumann, Mohit Iyyer, Matt Gardner, Christopher Clark,
  Kenton Lee, and Luke Zettlemoyer,
\newblock ``Deep contextualized word representations,''
\newblock in {\em NAACL}, 2018.

\bibitem{kusner2015doc}
Matt Kusner, Yu~Sun, Nicholas Kolkin, and Kilian Weinberger,
\newblock ``From word embeddings to document distances,''
\newblock in {\em ICML}, 2015.

\end{thebibliography}

\end{document}